\documentclass[runningheads]{llncs}

\usepackage{eccvabbrv}

\usepackage{graphicx}
\usepackage{booktabs}

\usepackage[accsupp]{axessibility}  

\usepackage{cite}
\usepackage{wrapfig}
\usepackage{enumitem}
\usepackage{multirow}
\usepackage{subcaption}
\usepackage{bm}
\usepackage{tikz}
\usepackage{pifont}
\usepackage[utf8]{inputenc}

\newcommand{\sref}[1]{\textsuperscript{\ref{#1}}}
\newcommand{\sblk}{\textsuperscript{ }}

\DeclareMathOperator*{\argmax}{arg\,max}

\usepackage[pagebackref,breaklinks,colorlinks]{hyperref}

\usepackage{orcidlink}

\begin{document}

\title{Large-Scale Multi-Hypotheses Cell Tracking \\ Using Ultrametric Contours Maps} 

\titlerunning{Large-Scale Cell Tracking Using UCMs}

\author{
Jord\~ao Bragantini\orcidlink{0000-0001-7652-2735} \and
Merlin Lange\orcidlink{0000-0003-0534-4374} \and
Lo\"ic Royer\orcidlink{0000-0002-9991-9724}
}

\authorrunning{J.~Bragantini \etal}

\institute{Chan Zuckerberg Biohub, San Francisco, USA
\email{\{jordao.bragantini,merlin.lange,loic.royer\}@czbiohub.org}}

\maketitle

\begin{abstract}

In this work, we describe a method for large-scale 3D cell-tracking through a segmentation selection approach.
The proposed method is effective at tracking cells across large microscopy datasets on two fronts: (i) It can solve problems containing millions of segmentation instances in terabyte-scale 3D+t datasets; (ii) It achieves competitive results with or without deep learning, bypassing the requirement of 3D annotated data, that is scarce in the fluorescence microscopy field. The proposed method computes cell tracks and segments using a hierarchy of segmentation hypotheses and selects disjoint segments by maximizing the overlap between adjacent frames. 
We show that this method is the first to achieve state-of-the-art in both nuclei- and membrane-based cell tracking by evaluating it on the 2D epithelial cell benchmark and 3D images from the cell tracking challenge. Furthermore, it has a faster integer linear programming formulation, and the framework is flexible, supporting segmentations from individual off-the-shelf cell segmentation models or their combination as an ensemble.
The code is available as supplementary material.

\end{abstract}

\section{Introduction}
\label{s.intro}

Tracking cells in terabyte-scale time-lapse volumetric image datasets of developing organisms acquired with the latest generation of light-sheet microscopes~\cite{Huisken:2004:LightSheet, Keller:2008:LightSheet, Stelzer:2021:LightSheetReview} is still an open problem. This conceptual and practical challenge involves segmenting individual cells and reconstructing their trajectories and lineages across multiple rounds of cell divisions. The scale of the data hinders the application of computationally intensive algorithms as these datasets easily contain thousands to millions of segmentation instances. Moreover, the lack of 3D ground-truth annotations also limits the application and performance of deep learning (DL) and other machine learning methodologies.

Methods usually tackle the problem in two different steps: computing the segmentation (or detection) and linking the previously detected instances given an association score. These methods can employ off-the-shelf segmentation models~\cite{Schmidt:2018:Stardist, Stringer:2021:Cellpose, Ouyang:2022:BioImageModelZoo}, and might use association scores from hand-crafted features~\cite{Magnusson:2016:SegmentationNTracking}. However, segmentation mistakes undermine the association score, harming the final tracking result. Because state-of-the-art segmentation methods are based on deep learning and require hard-to-obtain annotated data, their accuracy is greatly impaired when annotations are not available.

More recent methods based on flow-field estimation ~\cite{Hayashida:2019:CellMotionEstimation, Hayashida:2020:MPM, Sugawara:2022:ELEPHANT, Malin:2022:SparseAnnotTracking, Hirsch:2022:TrackingStructSVM} circumvent these limitations by avoiding segmentation altogether and estimating the position and motion of cells between frames, relying only on a marker in the center of each cell and a link between the same cell and their divisions in adjacent frames. The association score is computed through the motion estimation. This kind of cell-centroid supervision requires much less effort than annotating volumetric segmentation instances.

Rather than avoiding segmentation, other approaches leverage the spatial information and compute the cell segments and their tracks jointly by modeling the cell nuclei as a mixture of Gaussians~\cite{Amat:2014:TGMM, Schiegg:2013:ConservationTracking}, computing multiple segmentation hypotheses using ellipsoid fitting on their contours~\cite{Turetken:2016:NetworkFlow}, or formulating the splitting and linking of regions (\ie cells) in the image as a graph-cut problem in space and time ~\cite{Jug:2016:Moral, Tan:2019:Efficientnet, Funke:2018:BenchmarkEpithelialCellTracking}. While these approaches provide an elegant framework for joint segmentation and tracking, they are much more computationally intensive when compared to flow-based methods.

In practice, cell segmentations are invaluable because it assists subsequent downstream image analysis steps such as track validation and quantitative analysis (\eg morphological features, image intensity). Thus, we propose a novel method with the following goals in mind: 
\begin{itemize} 
    \item The segmentation instances should be computed jointly with tracking to avoid dependence on the accuracy of the segmentation method and to leverage both spatial and temporal information.
    \item The method should be computationally efficient and support out-of-memory processing of multi-terabyte datasets.
    \item It should leverage existing pre-trained models or traditional image processing for segmentation while achieving reasonable performance when ground-truth labels are unavailable.
\end{itemize}

To accomplish this, we propose a new algorithm for joint segmentation and tracking. It builds upon the concept of hierarchical segmentation and a novel method for extracting disjoint segments from a hierarchy and, at the same time, formulating it as an integer linear program (ILP) that respects the biological constraints (\eg, cell divides into two).

We perform multiple experiments evaluating different use cases of the proposed method and validate its performance against state-of-the-art-methods using the cell tracking challenge~\cite{Ulman:2017:CellTrackingChallenge} and the epithelial cell benchmark~\cite{Funke:2018:BenchmarkEpithelialCellTracking}. Moreover, we show our method can scale to large multi-terabyte scale real-world microscopy datasets~\cite{Yang:2022:Daxi} with millions of cell instances.

\section{Background and Related Works}
\label{s.relatedworks}

\subsection{Hierarchical Segmentation}
\label{s.segmrelworks}

Hierarchical segmentation has previously been used in cell tracking~\cite{Turetken:2016:NetworkFlow} but in a very limited scenario. Here, we review a more general representation of hierarchies and contours and their application to image segmentation.

Flat image segmentation (\ie non-hierarchical) concerns partitioning an image into disjoint segments, while hierarchical segmentation produces nested regions. Let $V$ be our basic units, the image voxels, a hierarchy $\mathcal H = \{S_1, S_2, ..., S_n\}$ contains a sequence of segmentations, where every $S_i$ is a collection of disjoint non-empty sets and their union is $V$. The sequence is ordered, such that $S_{i - 1}$ is a refinement of $S_i$, so any element in $S_{i - 1}$ is a subset of an individual region of $S_{i}$. Therefore, at the finest level, $S_1$, each region contains a single unit, and at the coarsest, there is a single region $S_n = \{V\}$, the whole image. By assigning an increasing value $\lambda_i$ for each partition $S_i$, the hierarchy can be represented as a dendrogram.

The hierarchical segmentation cannot be easily represented with a set of discrete labels as done for flat segmentation. Instead, they are commonly represented using an Ultrametric Contour Maps (UCM)~\cite{Najman:1996:Watershed, Najman:2011:EquivalenceWatershed, Arbelaez:2006:UCM} where the boundary between different regions receives the value $\lambda_i$ of where their merging happened.

Several methodologies explore this duality between fuzzy contour maps (\ie UCM, edge detection, affinity map) and hierarchies. Notably, Pont-Tuset~\etal~\cite{Pont:2016:MCG} accelerated the computation of normalized cut~\cite{Shi:2000:Normalized} of affinity maps by starting the clustering from a lower-resolution image and increasing its resolution, and finally aligning and merging the results into a UCM. Maninis~\etal~\cite{Maninis:2017:COB} used a convolutional neural network (CNN) to estimate boundaries of different orientations and applied the oriented watershed transform~\cite{Arbelaez:2010:OrientedWatershed} to obtain the hierarchical segmentation.

In the more general context, these nested segmentations are a hierarchical clustering of an edge-weighted graph $G = (V, E)$, where $V$ are the image voxels and $E$ the edges connecting nearby elements in the image planar lattice. For example, the watershed framework from~\cite{Najman:1996:Watershed} was extended to edge-weighted \emph{image graphs}~\cite{Cousty:2009:WaterCutMSF, Najman:2011:EquivalenceWatershed}.

In the microscopy domain, segmentation from edge weights is extremely popular~\cite{Briggman:2009:MALIS, Funke:2018:MALA, Wolny:2020:PlantSeg}. Because most applications expect a single label per instance, methodologies based on correlation clustering~\cite{Kappes:2011:GloballyOptimalSegmentationMulticuts, Keuper:2015:LiftedMulticuts, Yarkony:2012:FastPlanarCorrClustering, Pape:2017:LargeMultiCut} are preferred over hierarchical~\cite{Nunez:2013:GALA} because they avoid the extremely challenging step of selecting the segment from the hierarchy without using a naive horizontal cut in the dendrogram. Kiran and Serra~\cite{Kiran:2014:GlobalHierarchicalCut} proposed an algorithm that finds the optimum flat partition in a hierarchy for a restricted set of energies. More recently, GASP~\cite{Bailoni:2022:GASP} was proposed to unite correlation and hierarchical clustering.

Our approach is not constrained to any hierarchical segmentation method; we chose the watershed framework due to its simplicity and computational performance.

\subsection{Cell Tracking}
\label{ss.track}

Cell tracking has been a central problem for the bioimage community for more than two decades~\cite{Zimmer:2002:ActiveContoursTracking}. Several different strategies have been explored, a non-exhaustive list being: contour evolution~\cite{Dzyubachyk:2010:LevelSetTracking}; sequential fitting of mixtures of Gaussians~\cite{Amat:2014:TGMM}; dynamic programming using the Viterbi algorithm~\cite{Magnusson:2016:SegmentationNTracking}; and probabilistic graphical models that account for under/over-segmentation errors~\cite{Schiegg:2013:ConservationTracking}.

Most related to our approach are joint segmentation and tracking methods, where the final cell segments are found during the tracking step~\cite{Bise:2013:TrackingHighConfluency}. In~\cite{Turetken:2016:NetworkFlow}, the authors compute a hierarchical segmentation by fitting ellipsoids to their cells detection binary map contours, then, they estimate their association score using gradient boosting~\cite{Friedman:2002:StochasticGradientBoosting}; and obtain the tracks by solving a network flow ILP.

Jug~\etal~\cite{Jug:2016:Moral} proposed (and later improved~\cite{Rempfler:2017:EfficientMoral}) the moral lineage tracing (MLT) problem by formulating tracking as a minimum cost multi-cut subject to biological constraints: cells can divide but not merge, and that they can appear or disappear from the field of view. This approach extended the wide literature of multi-cut problems in the spatial domain (\ie, image segmentation)~\cite{Kappes:2011:GloballyOptimalSegmentationMulticuts, Keuper:2015:LiftedMulticuts, Yarkony:2012:FastPlanarCorrClustering, Pape:2017:LargeMultiCut} to include the time axis and additional constraints. Despite the elegant framework, the NP-hardness of the multi-cut problem limits its application to medium-size datasets, even with specialized algorithms~\cite{Rempfler:2017:EfficientMoral}.

More recently, flow-based methods~\cite{Hayashida:2019:CellMotionEstimation, Nishimura:2020:WeaklyCellTracing, Sugawara:2022:ELEPHANT} were proposed and have shown the most competitive performance. In~\cite{Hayashida:2020:MPM}, they estimate both the centroid detection and the motions of cells using a CNN. The tracking is computed by advecting backward in time using motion estimation and starting from the cells' detections.
Malin-Mayor~\etal~\cite{Malin:2022:SparseAnnotTracking, Hirsch:2022:TrackingStructSVM} extends this method by proposing a loss function that allows sparse annotation of the cells' centroids and their links, additionally, the lineages are obtained with an ILP solver rather than a heuristic as in earlier work~\cite{Hayashida:2019:CellMotionEstimation, Hayashida:2020:MPM, Sugawara:2022:ELEPHANT}.

Other notable DL-learning-based tracking methods employ different strategies:
\cite{Payer:2018:RecurrentHourglassTracking} tracks cells by clustering pixels from adjacent frames using their coordinates and embeddings from a recurrent hourglass network. Starting from a set of segments, Ben~\etal~\cite{Ben:2022:GraphNNCellTracking} construct a directed graph and solves the tracking using a graph neural network to classify the edges that belong to the cell lineages.

\section{Method}
\label{s.method}

Our method was developed for fluorescence microscopy images, where cells are engineered to express localized fluorescence, making them or their boundaries easily identifiable from the background. The challenge is to split the cells' instances, which cannot be easily resolved because of optical artifacts, and link these instances through time.

To avoid committing to a single potentially erroneous candidate segment per cell, we construct a hierarchy of segments. Ideally, the true cell instance will be one of these candidate segments. This differs from methods that aggregate superpixels during tracking~\cite{Jug:2016:Moral, Rempfler:2017:EfficientMoral} because the set of possible segments is fixed during the ILP optimization.

While it is rather optimistic to assume that the hierarchy will always closely fit the true segments, this approach can always find a single region at a finer level that belongs to a unique true segment at the cost of not covering it completely.

We formulate the problem of finding the instances in the hierarchy as an optimization problem to maximize the overlap between the \textit{unknown} true instances and a selection of disjoint segments from the hierarchy, which we will call maximum total intersection over union (MTIoU).

Given a hierarchy $\mathcal H$ and a set of ground-truth segments $\mathcal G = \{g_1, g_2, ..., g_n\}$, the set of disjoint segmentation with MTIoU can be found with the following ILP:
\begin{align}
    \argmax_{\mathbf x} &= \sum_{p \in \mathcal H} \sum_{q \in \mathcal G} w_{pq} x_{pq} \\
    \text{s.t.} \quad
    y_p &= \sum_{q \in \mathcal G } x_{pq} \quad \forall \ p \in \mathcal H \label{eq.sumh} \\
    \sum_{p \in \mathcal H } x_{pq} &\leq 1 \quad \forall \ q \in \mathcal G \label{eq.sumg} \\ 
     y_p + y_q &\leq 1 \quad \forall \ p, q \in \mathcal H | p \subset q \label{eq.Goverlap} \\
     y_p &\in \{0, 1\} \quad \forall \ p \in \mathcal H \\
     x_{pq} &\in \{0, 1\} \quad \forall \ p, q \in \mathcal H \times \mathcal G
\end{align}
where $w_{pq}$ is the intersection over union (IoU) between segment $p$ and $q$; $x_{pq}$ indicates if $p$ is assigned to $q$; $y_p$ indicates if segment $p$ was selected ($y_p = 1$) with Eq.~\ref{eq.sumh} forcing a single assignment per segment; Eq.~\ref{eq.sumg} forces a single assignment per ground-truth segment; Eq.~\ref{eq.Goverlap} restricts the solution to only disjoint segments (\ie they belong to different branches of the hierarchy).

In practice, information about the true instances is not available, so we approximate this problem by solving the segmentation selection between hierarchies of adjacent frames under the assumption that they are similar. Thus, the cell segments will overlap in some nodes in the hierarchies.

By formulating the ILP considering every time point, not just a pair of frames, and including the biological constraints, such as cell division, disappearing and appearing cells~\cite{Turetken:2016:NetworkFlow}, the resulting segmentation and their association between hierarchies can reconstruct the cell lineages.

Therefore, the ILP formulation of the MTIoU between hierarchies for cell tracking becomes:
\begin{align}
    \argmax_{\mathbf x} &= \sum_{t \in [2 .. T - 1]} \sum_{p \in \mathcal H_{t - 1}} \sum_{q \in \mathcal H_t} w_{pq} x_{pq} + \nonumber \\
    & \sum_{p \in \bm{\mathcal{H}} } w_{\alpha} x_{\alpha p} + w_{\beta} x_{p \beta} + w_{\delta} x_{\delta p} \label{eq.ilp} \\
    \text{s.t.} \quad
    y_q &= x_{\alpha q} + \sum_{p \in H_{t - 1}} x_{pq} \ \ \forall \ q \in \mathcal H_t \ \forall \ t \in [1 .. T] \label{eq.inflow} \\
    y_p + x_{\delta p} &= x_{p \beta} + \sum_{q \in H_{t + 1}} x_{pq} \ \ \forall \ p \in \mathcal H_t \ \forall \ t \in [1 .. T] \label{eq.outflow} \\
    y_p &\geq x_{\delta p} \ \ \forall \ \ p \in \bm {\mathcal H} \label{eq.div} \\
    y_p + y_q &\leq 1  \quad \forall \ p, q \in \mathcal H_t \ |  \ p \subset q \quad \forall \ t \in [1 .. T] \label{eq.overlap} \\
    y_p &\in \{0, 1\} \ \ \forall \ p \in \bm {\mathcal H} \label{eq.slacky} \\
    x_{pq} &\in \{0, 1\} \ \ \forall \ p, q \in \mathcal H_{t-1} \times \mathcal H_t \ \forall \ t \in [2 .. T] \label{eq.x}
\end{align}
where $\mathcal H_t$ is the candidate segments (\ie hierarchy) of frame $t$ and $\bm{\mathcal{H}} = \cup_{t \in [1 .. T]} \mathcal H _t$. We consider $\mathcal H_t = \emptyset$ when $t$ is out of range $[1 .. T]$; $x_{\alpha p}, x_{p \beta}, x_{\delta p}$ are the segment $p$ appearance, disappearance, and division binary variables, and their non-positive penalization parameters $w_\alpha, w_\beta, w_\delta$, respectively. For segments from $t = 1$, we set $w_\alpha = 0$, and $w_\beta = 0$ for $t = T$ as boundary conditions; $y_p$ is a slack binary variable indicating if segment $p$ is part of the solution. Equation~\ref{eq.inflow} and~\ref{eq.outflow} are the constant flow constraints~\cite{Turetken:2016:NetworkFlow} such that for the selected segments, there is an incoming segment from the previous frame or it is an appearing segment, and the outflow allows up to 2 out-going segments (\ie division) or its disappearance; Eq.~\ref{eq.div} asserts there are only divisions from an existing cell; Eq.~\ref{eq.overlap} restricts the solution to disjoint segments as in Eq.~\ref{eq.Goverlap}. In Section~\ref{ss.ilp}, we compare the run time of our ILP formulation with previous work.


\subsection{Hierarchy construction}

While not limited to any hierarchical segmentation method, we chose to compute the hierarchies using the hierarchical watershed framework~\cite{Perret:2017:EvaluationHierWS, Perret:2019:Higra}, which allows constructing hierarchies in log-linear time (sorting complexity) with respect to the number of edges in a graph with the minimum-spanning tree algorithm~\cite{Najman:2013:PlayingKrusKal} and Tarjan union-find~\cite{Tarjan:1975:UnionFind}.

For each frame $t$, we estimate a set of candidate segments, $\mathcal H_t$, from a binary foreground, $F_t$, and a fuzzy contour map, $C_t$, provided by the user. The foreground $F_t$ indicates the location of cells against the background. The contour map indicates where the cell boundaries (\ie contours) may lie. Since the connected components in $F_t$ are disconnected by definition, a hierarchy is computed for each one and integrated into a single $\mathcal H_t$ while ignoring the background.

Each hierarchy is constructed using the watershed hierarchy by area~\cite{Meyer:1997:MorphologicalTools, Perret:2019:Higra} on the undirected weighted graph from their respective connected component in $F_t$. The edge weight between a pair of pixels is their average intensity.

Using the full hierarchies on the tracking ILP (Eq.~\ref{eq.ilp}) can easily make the problem too large to solve. Therefore, three criteria were used to filter irrelevant segments~\cite{Perret:2019:RemovingRegions} and decrease the number of candidate solutions:
\begin{enumerate}
    \item Minimum size: the hierarchy is pruned below a minimum size threshold;
    \item Maximum size: the hierarchy is cut above a maximum size threshold;
    \item Non-relevant contours: segments with average contour strength below a threshold are fused.
\end{enumerate}

\begin{figure}
    \centering
    \begin{subfigure}[b]{0.48\textwidth}
        \centering
        \includegraphics[width=\textwidth]{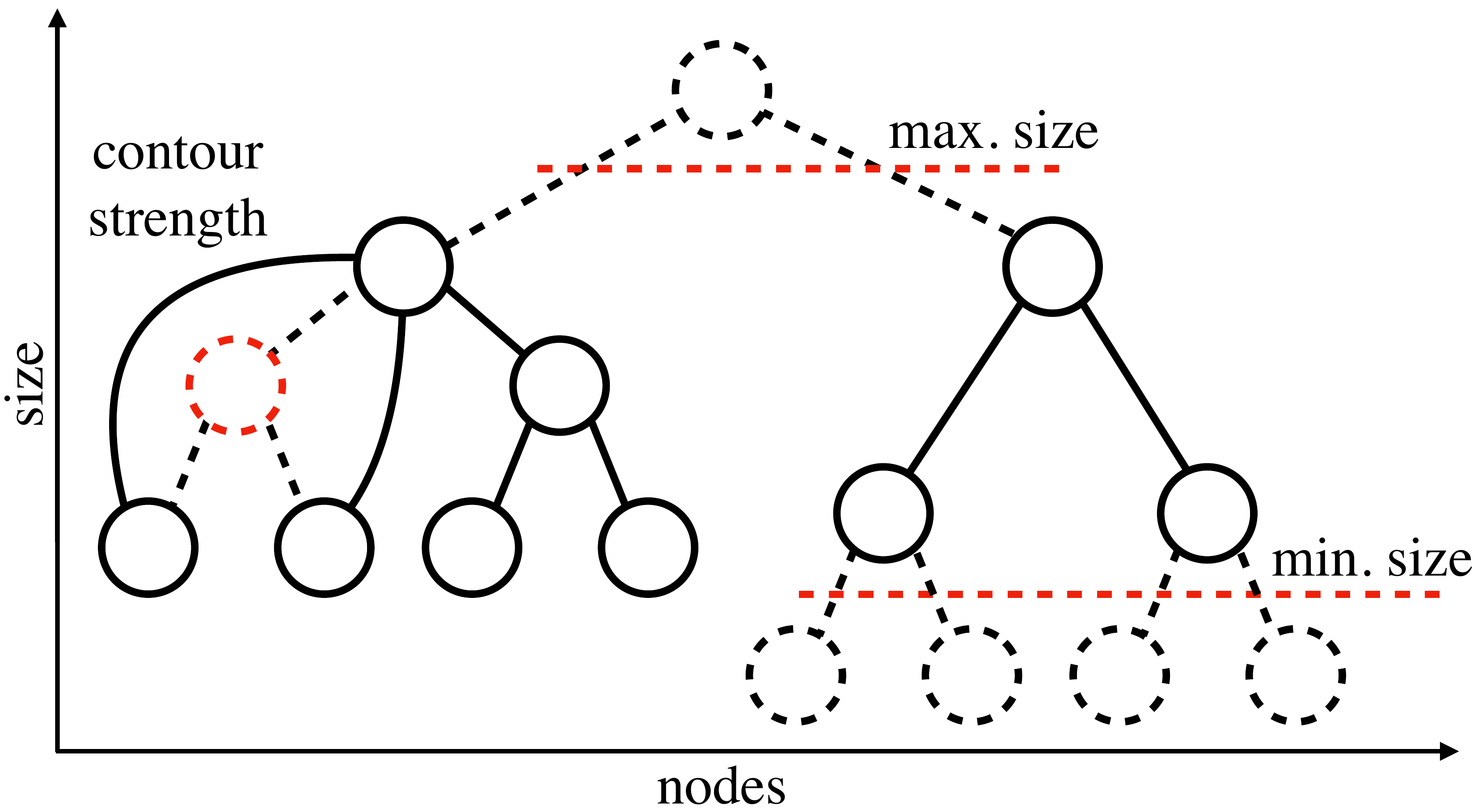}
        \caption{Example of hierarchy filtering by minimum size, maximum size, and contour strength; dashed lines show deleted elements. Leaves of the hierarchy are not shown}
        \label{f.hierfiltering}
    \end{subfigure}
    \hfill
    \begin{subfigure}[b]{0.48\textwidth}
        \centering
        \includegraphics[width=\textwidth]{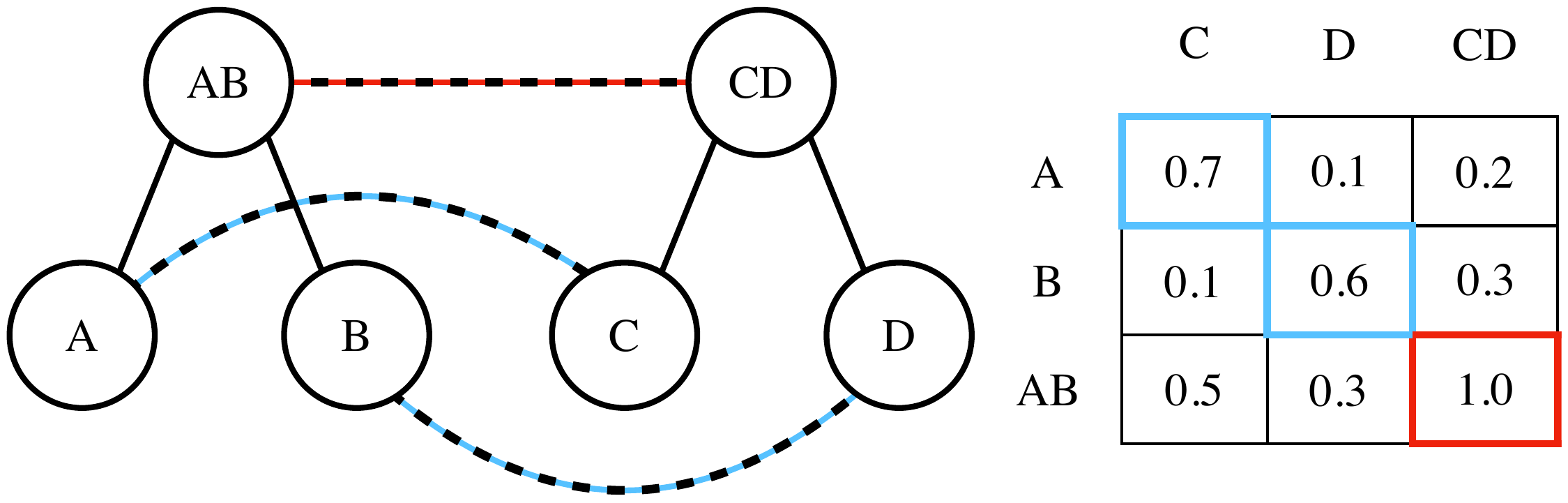}
        \caption{Example of linkage between two hierarchies and their weights matrix $w_{..}$ which includes all $w_{p,q}$. Solution considering $w_{..}$ in blue and in red solution of $w^p_{..}$ when $p >= 2$}
        \label{f.ioupower}
        \vspace{0.5cm}
    \end{subfigure}
\end{figure}

The minimum and maximum cell sizes are usually known for each organism, and the non-relevant contours filtering can be tuned by observing the under (over) segmentation of the solution. Figure~\ref{f.hierfiltering} shows a hierarchy filtering example.

The MTIoU can suffer from over-segmentation when spurious segments with high overlap between frames are present in the hierarchies. For example, in Figure~\ref{f.ioupower}, despite $AB$ and $CD$ having perfect overlap ($w_{AB,CD} = 1.0$) when solving Eq.~\ref{eq.ilp}, the solution selects segments $A, B, C, D$ (blue solution) because their sum of weights is larger. By taking a power $p$ of $w_{..}$, we reduce the over-segmentation as it favors fewer segments with larger $w_{..}$ values (red solution) rather than many with smaller $w_{..}$ values.

\subsection{Scaling to terabyte-scale datasets}
\label{ss.scaling}

As described previously, one major challenge of joint segmentation and tracking is the intense computational requirements. This section will discuss how our approach can effectively scale to terabyte-scale datasets.

\textbf{The hierarchical segmentation} is computed per connected component of each foreground map $F_t$, greatly reducing the number of pixels considered. For each hierarchy, the segments outside of the minimum and maximum size range are removed. Next, the disjoint segmentation constraint, Eq.~\ref{eq.overlap}, is computed by going through each segment to its last ancestor.
For each segment, we store their center coordinates and bounding boxes for computing their IoU.
Since the segmentation of each volume (plane) is independent, it can be easily run in parallel.

\textbf{The association weight between segments} is only computed for a subset of $k \times N_t$ pairs, where $N_t$ is the total number of segments on frame $t$. For $(t - 1, t)$-pair of frames, we start by computing a KDTree~\cite{Virtanen:2020:Scipy} with the segments' coordinates from $t - 1$ and querying the $2k$-nearest neighbors within a certain radius for each segment coordinate in $t$. From these $2k$ samples, we select the $k$ with the highest IoU, thus, we avoid computing the $w_{..}$ for every pair in $(t - 1, t)$, as $w_{..}$ is significantly sparse.
Additionally, since the IoU only considers the non-zero values of the segmentation masks, it can be computed using a crop around the segments' bounding box. When the bounding boxes between a segment and its neighbor do not intersect, its IoU is zero and does not require further computation.
The weights between different $(t - 1, t)$ are independent, so each pair of frames can be run in parallel.

\textbf{The ILP} can easily reach millions of variables due to multiple candidate segments per cell. When that happens, it cannot be solved in a reasonable amount of memory or time. However, its global solution can be approximated by solving the problem in multiple overlapping time windows as in~\cite{Malin:2022:SparseAnnotTracking}. The regions between overlapping windows are not used in the final solution except for a single frame, where the segments from already computed solutions are used as constraints, enforcing continuity of tracks, essentially stitching adjacent windows.

Malin-Mayor~\etal~\cite{Malin:2022:SparseAnnotTracking} employed their own distributed computing scheduler to process these windows in parallel while avoiding concurrent processing of overlapping windows. We propose a simpler two-pass interleaved scheduling scheme: i) Split the timelapse only along the time-axis into $N$ overlapping windows, where the overlap is smaller than a single window; ii) Solve the ILPs of the even-numbered window; iii) Add the solutions from the even-numbered ILPs as constraints; and iv) solve the odd-numbered ILPs. This simple strategy guarantees that no overlapping window is processed simultaneously, and it can be trivially executed in parallel at the cost of waiting for every even-numbered window to finish before starting the odd-numbered executions. Further improvements can be made by waiting for only adjacent pairs to finish (\eg window 3 depends on 2 and 4).

\section{Experiments}
\label{s.experiments}

We evaluate our claims and the proposed method in four different experiments:
\begin{enumerate}
    \item Multiple quantitative comparisons of the proposed algorithm against state-of-the-art methods for cell tracking.
    \item Analysis of an ensemble of segmentations with traditional segmentation algorithms and off-the-shelf segmentation CNNs.
    \item Runtime comparison of different ILP formulations.
    \item A multi-terabyte dataset cell tracking use case.
\end{enumerate}

When available, the experiments were evaluated using the metrics proposed by the Cell Tracking Challenge (CTC)~\cite{Ulman:2017:CellTrackingChallenge}: TRA, a score of operations required to transform the resulting lineage graph to the ground-truth lineage~\cite{Matula:2015:AOGM}; SEG, the average IoU between the predicted and the ground-truth segmentation instances (a predicted instance is assigned to a ground-truth instance when it overlaps with more than half of its pixels); and CTB, the average between the TRA and SEG scores, this is the primary ranking score in the competition and used to characterize state-of-the-art method.

\subsection{Quantitative comparison}
\label{ss.ctcexp}

\textbf{Cell Tracking Challenge:} This first quantitative experiment was run on five datasets from CTC: Fluo-C3DL-MDA231: Infected human breast carcinoma cells; Fluo-N3DH-CE: C. Elegans developing embryo~\cite{Murray:2008:CElegans}; Fluo-N3DL-DRO: D. Melanogaster developing embryo~\cite{Amat:2014:TGMM}; Fluo-N3DL-TRIF (TRIC): Developing Tribolium Castaneum embryo (and its cartographical projection). In each case, two timelapses are available for training and validation purposes, and two other timelapses are used for testing. Test set results were evaluated by the competition organizers, and their labels are not publicly available.

The MDA231 and CE datasets have ground-truth tracks and silver-standard segmentation annotation automatically generated by a consensus~\cite{Akbacs:2019:AutomaticFusion} of the top-ranking algorithms from previous iterations of CTC. The DRO, TRIF, and TRIC datasets only have ground-truth tracks for a specific group of cells and a few 2D labels. In these cases where not every lineage are labeled, the initial positions of these cells' tracks are provided during submission and must be used to select only the cells that belong to these lineages. These sparse annotations make the tracking much more challenging because less training data are available, and during evaluation, other lineages must be removed, so missing links on the predicted lineages will prune the rest of the tracks and greatly decrease the final score.

\begin{minipage}[t]{.46\textwidth}
\centering
\captionof{table}{
    Results on the hidden test dataset of the cell tracking challenge at the time of article submission;
    subscripts indicate the method used according to enumeration; our submission is in bold (anonymous submission on the competition website until paper acceptance)
}
\small
\begin{tabular}{@{}lccccc@{}} \toprule
Dataset                   & Rank & CTB~\sblk            & TRA~\sblk            & SEG~\sblk           \\ \midrule
\multirow{4}{*}{MDA}   & 1st  & 0.797~\sref{en.kit}  & 0.884~\sref{en.kit}  & 0.710~\sref{en.kit} \\
                          & 2nd  & 0.761~\sref{en.leid} & 0.882~\sref{en.kth}  & 0.642~\sref{en.leid}\\
                          & 3rd  & 0.757~\sref{en.kth}  & 0.880~\sref{en.leid} & 0.632~\sref{en.kth} \\
                          & 7th  & \textbf{0.683}~\sblk & \textbf{0.818}~\sblk & \textbf{0.549}~\sblk\\ \midrule
\multirow{4}{*}{CE}       & 1st  & \textbf{0.844}~\sblk & 0.987~\sref{en.mpi}  & 0.759~\sref{en.mu}  \\
                          & 2nd  & 0.829~\sref{en.kth}  & 0.979~\sref{en.jan}  & 0.729~\sref{en.kit} \\
                          & 3rd  & 0.808~\sref{en.kit}  & 0.975~\sref{en.igfl} & 0.722~\sref{en.kth} \\
                          & 4th  & -                    & \textbf{0.967}~\sblk & \textbf{0.722}~\sblk\\ \midrule
\multirow{3}{*}{DRO}      & 1st  & \textbf{0.708}~\sblk & \textbf{0.802}~\sblk & \textbf{0.613}~\sblk \\
                          & 2nd  & 0.617~\sref{en.kth}  & 0.785~\sref{en.jan}  & 0.613~\sref{en.kth} \\
                          & 3rd  & 0.591~\sref{en.jan}  & 0.785~\sref{en.mpi}  & 0.397~\sref{en.jan} \\ \midrule
\multirow{3}{*}{TRIF}     & 1st  & \textbf{0.841}~\sblk & 0.955~\sref{en.mpi}  & \textbf{0.746}~\sblk\\
                          & 2nd  & 0.804~\sref{en.mpi}  & \textbf{0.936}~\sblk & 0.684~\sref{en.rwth}\\
                          & 3rd  & 0.785~\sref{en.rwth} & 0.886~\sref{en.rwth} & 0.654~\sref{en.mpi} \\ \midrule
\multirow{4}{*}{TRIC}     & 1st  & 0.867~\sref{en.kth}  & 0.952~\sref{en.mpi}  & 0.791~\sref{en.kth} \\
                          & 2nd  & 0.816~\sref{en.mpi}  & 0.942~\sref{en.kth}  & 0.766~\sref{en.kit} \\
                          & 3rd  & 0.787~\sref{en.kit}  & \textbf{0.854}~\sblk & 0.680~\sref{en.mpi} \\
                          & 4th  & \textbf{0.754}~\sblk & -                    & \textbf{0.654}~\sblk\\ \bottomrule
\end{tabular}
\label{t.test}
\end{minipage}
\hfill
\begin{minipage}[t]{0.48\textwidth}
\centering
\captionof{table}{Epithetial Cell Benchmark segmentation (SEG metric) results, top scores in bold, higher is better. Our tracking algorithm is indicated by \ding{61}}
\small
\begin{tabular}{@{}cc@{}c@{}cc@{}cc@{}} \toprule
Method && Variation                                       && PER                        && PRO \\ \cmidrule{1-1} \cmidrule{3-3} \cmidrule{5-5} \cmidrule{7-7}
FFN    && Default                                         && $0.879$          && $0.796$ \\
MTL    && Default                                         && $0.904$          && $\mathbf{0.818}$ \\
TA     && Default                                         && -                && $0.758$ \\ \cmidrule{1-1} \cmidrule{3-3} \cmidrule{5-5} \cmidrule{7-7} 
\multirow{2}{*}{PlantSeg} && Default model                && $0.787$          && $0.761$ \\
                          && Fine tuned                   && $0.885$          && $0.800$ \\ \cmidrule{1-1} \cmidrule{3-3} \cmidrule{5-5} \cmidrule{7-7}
\multirow{3}{*}{MALA}     && Original                     && $\mathbf{0.907}$ && $\mathbf{0.817}$ \\
                          && Our repeat                   && $0.902$          && $0.810$ \\
                          && Our tracking$^\text{\ding{61}}$  && $\mathbf{0.909}$ && $\mathbf{0.817}$ \\  \cmidrule{1-1} \cmidrule{3-3} \cmidrule{5-5} \cmidrule{7-7}
Blur                      && Our tracking$^\text{\ding{61}}$  && $0.853$          && $0.804$ \\ \bottomrule
\end{tabular}
\label{t.dro}
\centering
\captionof{table}{Tracking and segmentation results for individual segmentation methods and their ensemble}
\begin{tabular}{@{}clcccccc@{}} \toprule
File               & Method    &&  TRA   &  SEG   & CTB \\ \midrule
\multirow{4}{*}{1} & Watershed && 0.905  & 0.593  & 0.749 \\
                   & Cellpose  && 0.863  & 0.671  & 0.767 \\
                   & Stardist  && 0.910  & 0.724  & 0.817 \\ \cmidrule{2-2} \cmidrule{4-6}
                   & Ensemble  && \textbf{0.956}  & \textbf{0.740}  & \textbf{0.848} \\ \midrule
\multirow{4}{*}{2} & Watershed && 0.898  & 0.695  & 0.796 \\
                   & Cellpose  && 0.908  & 0.743  & 0.825 \\
                   & Stardist  && 0.941  & 0.818  & 0.880 \\ \cmidrule{2-2} \cmidrule{4-6}
                   & Ensemble  && \textbf{0.956}  & \textbf{0.826}  & \textbf{0.891} \\ \bottomrule
\end{tabular}
\label{t.ensemble}
\end{minipage}
\vspace{0.5cm}

We used a U-Net CNN~\cite{Ronneberger:2015:Unet} for predicting the cell and edge detection maps, and both maps were optimized using their own dice loss~\cite{Deng:2018:CrispBoundaries} with additional dice losses on side edge-detection predictions~\cite{Xie:2015:Holistically} during training.

The DRO, TRIF, and TRIC do not have any 3D segmentation annotations. Therefore, we tried two approaches. i) Using our U-Net and training it using labels computed from a traditional segmentation algorithm. We denoted this as pseudo-labels (PL). ii) Using filters to detect the cells and compute the contour topology, denoted by image processing (IM).

In both cases, the foreground (\ie cells) was detected using a difference of Gaussians and thresholding it with Otsu. The contour maps were simply the inverse of the Gaussian blurred image. For the pseudo-labels, we used the contour maps and foregrounds to run the hierarchical watershed and performed a horizontal cut in the hierarchy to obtain the flat segments used for training our CNN. In IM, the foreground and contour maps were used as the tracking input.

We optimized the hyper-parameters using grid search and cross-validation on the training set to avoid a biased estimate of our final accuracy, so models trained on dataset 01 were applied to dataset 02 and the other way around. More details about these procedures, including their parameters, can be found in the supplementary materials.

The non-deep-learning solutions achieved better results (see supplementary) on these datasets than deep learning combined with pseudo-labels, therefore, IM was used in our final submission. Next, the proposed method was evaluated on the hidden test set.

The competition organizers report the top three methods for each metric and provide access to our ranking regardless of our position. The final results are reported in Table~\ref{t.test}.

The leaderboard's methods at the time of submission were:
\setlist{nolistsep}
\begin{enumerate}[noitemsep]
    \item KTH-SE:~\cite{Magnusson:2016:SegmentationNTracking} image processing solution with tracking using Viterbi algorithm; \label{en.kth}
    \item KIT-GE: EmbedTrack~\cite{Loffler:2022:EmbedTrack}, a joint flow-based tracking and pixel embedding segmentation method; \label{en.kit}
    \item MPI-GE-CBG: Description is not publicly available, source code indicates it is a DL-based method with nearest neighbor linking; \label{en.mpi}
    \item JAN-US: ~\cite{Hirsch:2022:TrackingStructSVM} it is an improved version of~\cite{Malin:2022:SparseAnnotTracking}, where they classify the cell states to detect division and assist the ILP formulation and use a structured SVM to find the optimal weights of the ILP automatically; \label{en.jan}
    \item LEID-NL: Model-evolution~\cite{Dzyubachyk:2010:LevelSetTracking} with routines to split dividing cells and track cells entering the field of view; \label{en.leid}
    \item MU-CZ: H-minima watershed~\cite{Meyer:2005:MorphSegmRevisited} plus U-Net for cell contour and foreground prediction, tracking uses an intersection-based heuristic; \label{en.mu}
    \item IGFL-FR: ELEPHANT~\cite{Sugawara:2022:ELEPHANT} uses the cell detection and advection of DL flow prediction linking as item~\ref{en.mpi}; \label{en.igfl}
    \item RWTH-GE:~\cite{Stegmaier:2014:TerabyscaleNucleiTracking} Laplacian of Gaussian cell detection, with nearest neighbor linking with manual curation for missing detections. \label{en.rwth}
\end{enumerate}

The large variety of top-ranking methods shows that there is not a single best-tracking algorithm. Despite this, our approach obtained the best performance in the combined score, CTB, on three out of the five datasets: CE with supervised DL and TRIF and DRO with image processing only. In the TRIC dataset, most of our mistakes were on the periphery of the projection, where the cells' movements are inflated, making our IoU weights significantly less effective. In the MDA231, we achieved our worst ranking. We suspect our edge-detection network did not generalize to the test dataset.

\textbf{Epithelial Cell Benchmark}~\cite{Funke:2018:BenchmarkEpithelialCellTracking} evaluates a different tracking modality, where the cells are densely packed, and the cells' membranes are visible rather than their nucleus. It contains eight 2D timelapses, divided into two cell types, three peripodial (PRE) and five proper discs (PRO) cells' timelapses~\cite{Aigouy:2016:DroEpithelial}. In this context, graph-cut methodologies extended from the electron microscopy segmentation domain are commonly used~\cite{Funke:2018:MALA, Wolny:2020:PlantSeg}.

We compared with the original baselines MTL~\cite{Rempfler:2017:EfficientMoral}, FFN~\cite{Januszewski:2018:FFN}, TA~\cite{Aigouy:2016:DroEpithelial}, MALA~\cite{Funke:2018:MALA} and PlantSeg~\cite{Wolny:2020:PlantSeg} which uses Multicut from ilastik~\cite{Berg:2019:ilastik} and GASP~\cite{Bailoni:2022:GASP}. Additionally, we reproduced MALA's experiment~\footnote{\small\url{https://github.com/funkey/flywing}}, achieving a score lower than the original run but still competitive, Table~\ref{t.dro}. Refer to~\cite{Funke:2018:BenchmarkEpithelialCellTracking} for a more in-depth description of baselines.

We evaluated our algorithm in two scenarios, marked with a \ding{61} \ in Table~\ref{t.dro}: (i) the original frames filtered with a Gaussian kernel ($\sigma = 1.0$) as our contour map, (ii) using MALA's edge predictions as input.

The minimal solution without deep learning (i) is comparable to some of the DL approaches, especially in the PRO datasets, which have diverse cell sizes. Our solution from MALA's boundaries (ii) meets or surpasses other methods despite having inferior predictions than the original MALA work. The performance could be further improved by accessing MALA's original network weights. This displays the effectiveness of our approach at performing the correct segmentation against state-of-the-art DL and graph segmentation schemes. Videos are attached as supplementary.

\subsection{Segmentation ensemble experiment}

Most biologists cannot train their segmentation models or design complex image-processing pipelines. Therefore, leveraging broadly available segmentation tools is essential. We show that our method can use segmentations from different algorithms and combine them to achieve superior performance using fuzzy contour map representations on the training set of Fluo-N2DL-HeLa~\cite{Neumann:2010:Fluo-N2DL-HeLa}, a 2D dataset of HeLa cells expressing H2b-GFP from CTC.
We chose this dataset since it contains high-quality images and annotations. Because it is a 2D dataset, we can leverage state-of-the-art off-the-shelf 2D segmentation methods such as Cellpose~\cite{Stringer:2021:Cellpose} and Stardist~\cite{Schmidt:2018:Stardist}.

\begin{figure}
    \centering
    \begin{subfigure}[b]{0.19\columnwidth}
        \centering
        \includegraphics[width=\textwidth]{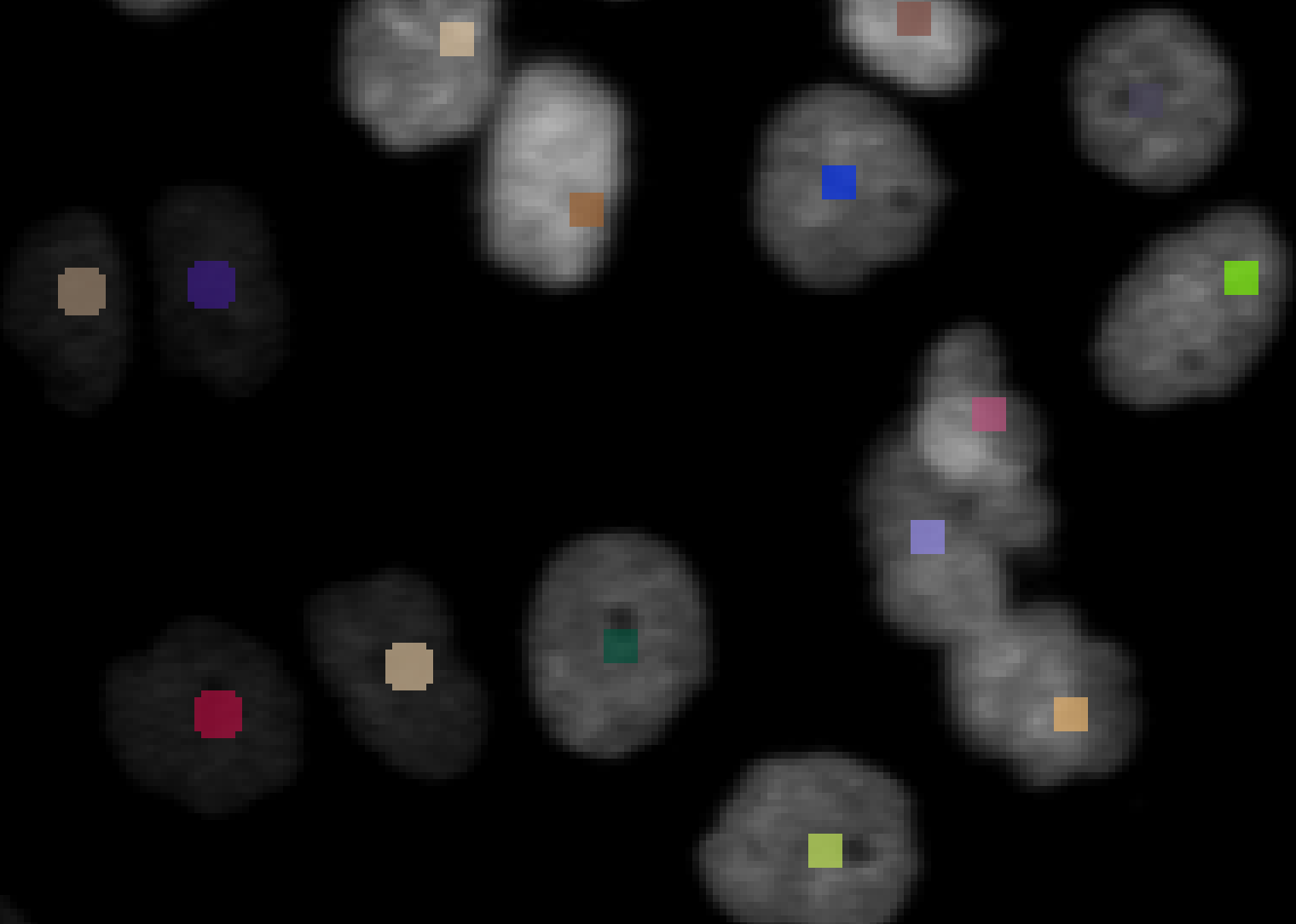}
        \caption{GT}
    \end{subfigure}
    \begin{subfigure}[b]{0.19\columnwidth}
        \centering
        \begin{tikzpicture}
            \node[anchor=south west,inner sep=0] {\includegraphics[width=\textwidth]{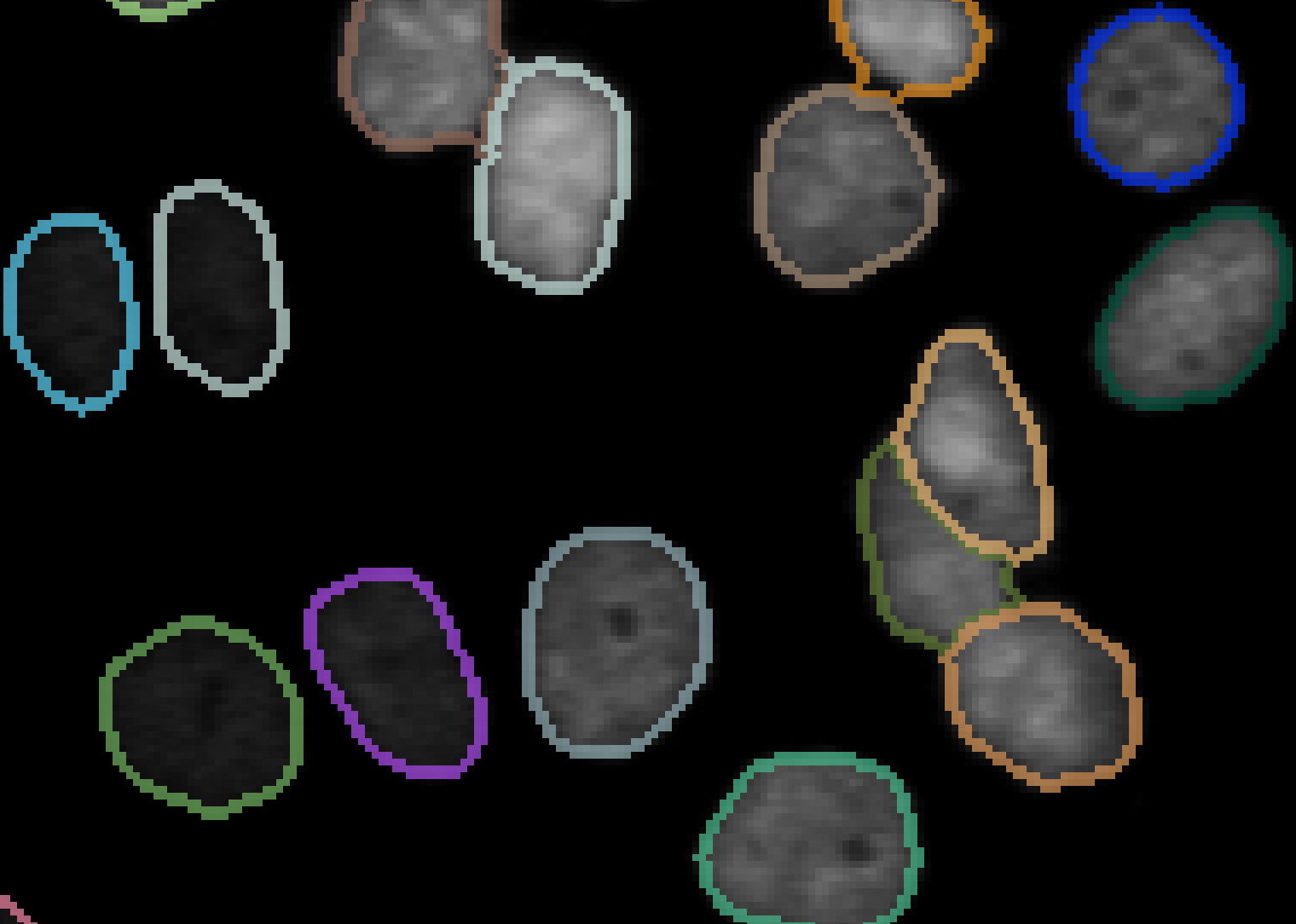}};
        \end{tikzpicture}
        \caption{ensemble}
    \end{subfigure}
    \begin{subfigure}[b]{0.19\columnwidth}
        \centering
        \begin{tikzpicture}
            \node[anchor=south west,inner sep=0] {\includegraphics[width=\textwidth]{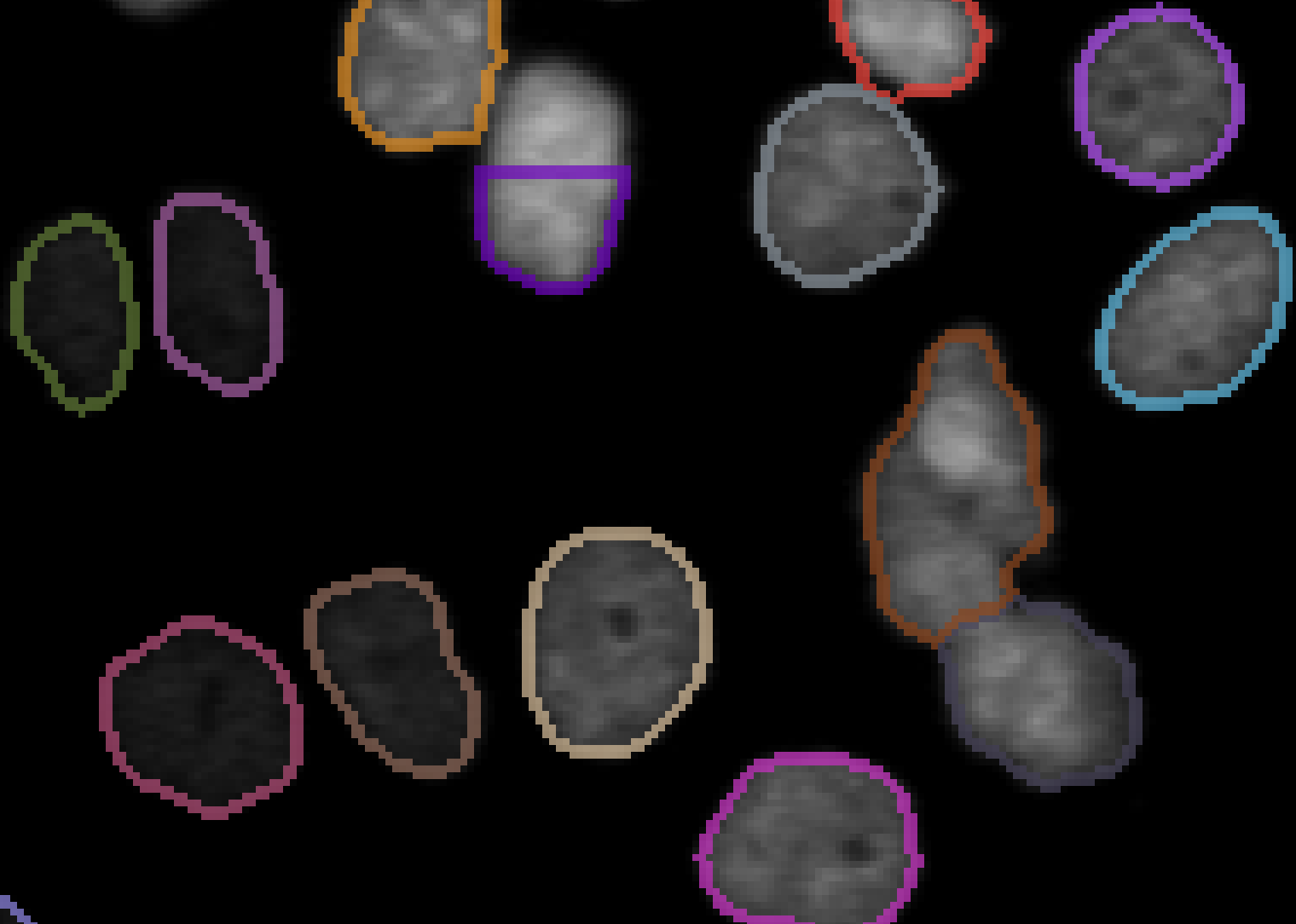}};
            \draw[-stealth, line width=2.5pt, red] (1.15,1.0) -- (1.55,0.7);
            \draw[-stealth, line width=2.5pt, red] (0.5,1.1) -- (0.9,1.3);
        \end{tikzpicture}
        \caption{watershed}
    \end{subfigure}
    \begin{subfigure}[b]{0.19\columnwidth}
        \centering
        \begin{tikzpicture}
            \node[anchor=south west,inner sep=0] {\includegraphics[width=\textwidth]{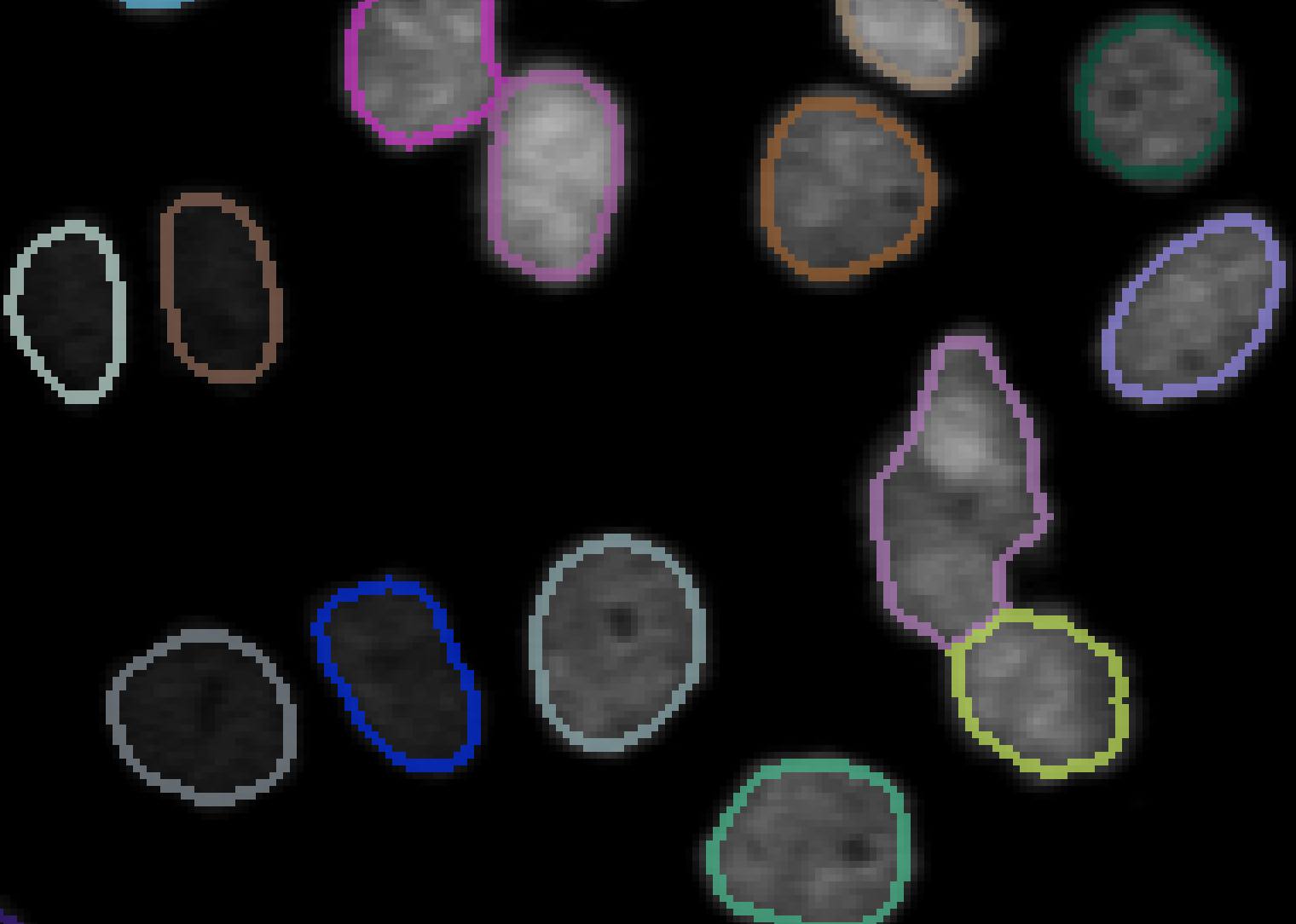}};
            \draw[-stealth, line width=2.5pt, red] (1.15,1.0) -- (1.55,0.7);
        \end{tikzpicture}
        \caption{cellpose}
    \end{subfigure}
    \begin{subfigure}[b]{0.19\columnwidth}
        \centering
        \begin{tikzpicture}
            \node[anchor=south west,inner sep=0] {\includegraphics[width=\textwidth]{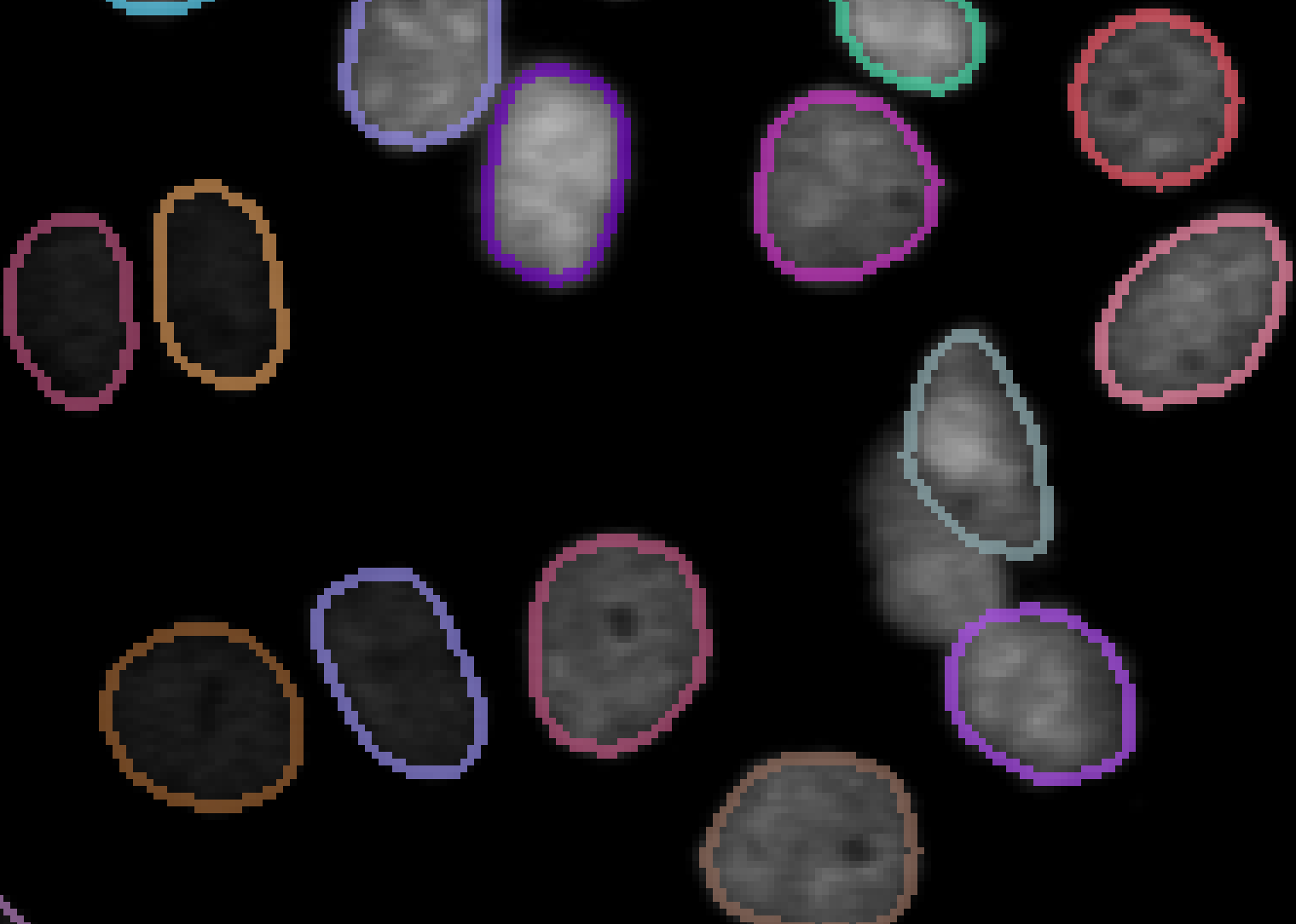}};
            \draw[-stealth, line width=2.5pt, red] (1.15,1.0) -- (1.55,0.7);
        \end{tikzpicture}
        \caption{stardist}
    \end{subfigure}
    \caption{Results from individual methods versus their ensemble, mistakes are indicated by the red arrows. (a) ground-truth cell detection; (b) result; (c-e) segmentation results of different methods}
    \label{f.ensemble}
\end{figure}

We evaluate three different segmentation methods: a classical watershed from h-minima using a distance transform as the watershed basins~\cite{Lotufo:2002:IFTGray, Van:2014:SKimage}; Cellpose~\cite{Stringer:2021:Cellpose} using their nuclei pre-trained model; and Stardist~\cite{Schmidt:2018:Stardist} using their "2d versatile fluorescence" pre-trained model.

For each segmentation map, we extract a binary map indicating foreground (\ie cell) versus background and the distinct labels' binary contours. The ensemble maps were created by combining the foregrounds using the logical OR operator and the contours by averaging.
Table~\ref{t.ensemble} presents the scores from running our proposed algorithm on the foreground and contour maps from the labels of each method. The top score for the ensemble model shows that the contour map representation is an effective way to combine multiple segmentation, and our ILP picks segments that are better than any individual segmentation result, consequently improving cell tracking. Figure~\ref{f.ensemble} shows how this approach benefits segmentation and tracking.

\subsection{ILP run time comparison}
\label{ss.ilp}

\begin{wrapfigure}{r}{7cm}
    \vspace{-1.25cm}
    \centering
    \includegraphics[width=7cm]{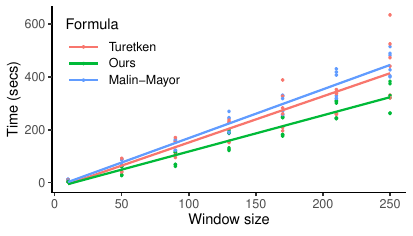}
    \caption{Runtime comparison between ILP formulations}
    \vspace{-0.75cm}
    \label{f.ilp.runtime}
\end{wrapfigure}

Cells moving in and out of the field of view and cell division constraints are critical aspects of the cell-tracking ILP and can be formulated differently. In this experiment, we compare the performance of our formula against other recent ILP-based methods for optimizing Eq.~\ref{eq.ilp}, T{\"u}retken~\etal~\cite{Turetken:2016:NetworkFlow} and Malin-Mayor~\etal\cite{Malin:2022:SparseAnnotTracking}, with the same experimental setup and parameters. We included Eq.~\ref{eq.overlap} to Malin-Mayor formulation to allow multiple segmentation hypotheses and set $w_\beta = 0$ for all methods because Malin-Mayor does not penalize disappearance events.

We compared the average run time of 3 runs with increasingly longer timelapses (\ie window) of the Fluo-N3DH-CE datasets. The window started from the middle of the timelapse and increased at both ends.
A linear fit for each method reported slope coefficients of $1.75\pm0.12$ for T{\"u}retken, $\mathbf{1.37\pm0.06}$ for Ours, and $1.84\pm0.06$ for Malin-Mayor, the smaller, the better.
Figure~\ref{f.ilp.runtime} display the measurements and their linear regression. Therefore, our formulation is faster and scales better with increasingly larger problems. All sizes except the initial size of 10 produce reasonable tracking results.

\subsection{Terabyte-scale tracking use case}
\label{ss.wholeembryo}

We ran our algorithm on a 3D time-lapse dataset of a developing zebrafish embryo~\cite{Yang:2022:Daxi} recorded with a light-sheet microscope. The recording starts at 12 hrs post-fertilization and lasts 6.6 hrs with an acquisition interval of 30 secs. After cropping and stabilizing the embryo region, the dataset's uncompressed size is about 3.4 TB, and its shape is $791 \times 448 \times 2174 \times 2423$ voxels (t, z, y, x).

The algorithm was executed in a compute cluster using SLURM~\cite{Yoo:2003:SLURM}, following the strategies described in Section~\ref{ss.scaling}. For the tracking step, a window of size 50 with an overlap of 5 frames on each side was used. For a fair comparison, we limited our resources to 100 CPU cores and 20 GPUs as used in~\cite{Malin:2022:SparseAnnotTracking}. 

Processing took approximately a total of 9.5 hours of distributed computing, including: 1 hour and 20 min. to detect the foreground and predict the contours using a U-Net as described in the other experiments, and trained on pseudo-labels from a watershed segmentation;  2 hours and 7 min. to compute the hierarchical segmentation for all frames; 40 minutes for the association between segments; 3 hours for the tracking first pass; 2 hours and 8 minutes for the second pass; and 18 minutes to export the data. The first step used GPUs, the remaining only CPU cores, processing in total 25 hours of GPU time and 31 days 22 hours of CPU time. Resulting in more than 21.5 million cell instances. The ILP size is partially proportional to the number of resulting instances.

Results are shown in Figure~\ref{f.wholeembryo} and in a video in supplementary materials. Annotations are not available for a quantitative analysis. However, we expect an accuracy similar to embryonic datasets from the cell tracking challenge, Tab.~\ref{t.test}.

In comparison, with similar resources, Malin-Mayor~\etal\cite{Malin:2022:SparseAnnotTracking} tracked a mouse dataset in an earlier developmental stage (therefore, fewer and less densely distributed nuclei) in 44 hours, segmenting about 7 million cell instances. 
Information about the total GPU and CPU time is not available. 

\begin{figure} 
    \centering
    \begin{subfigure}[b]{0.22\columnwidth}
        \includegraphics[width=\textwidth]{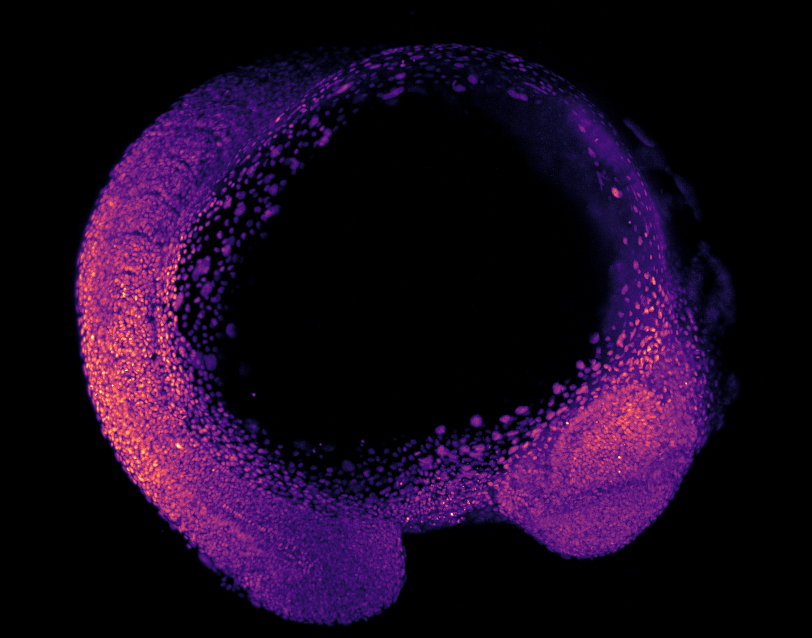}
        \caption{}
    \end{subfigure}
    \hfill
    \begin{subfigure}[b]{0.22\columnwidth}
        \includegraphics[width=\textwidth]{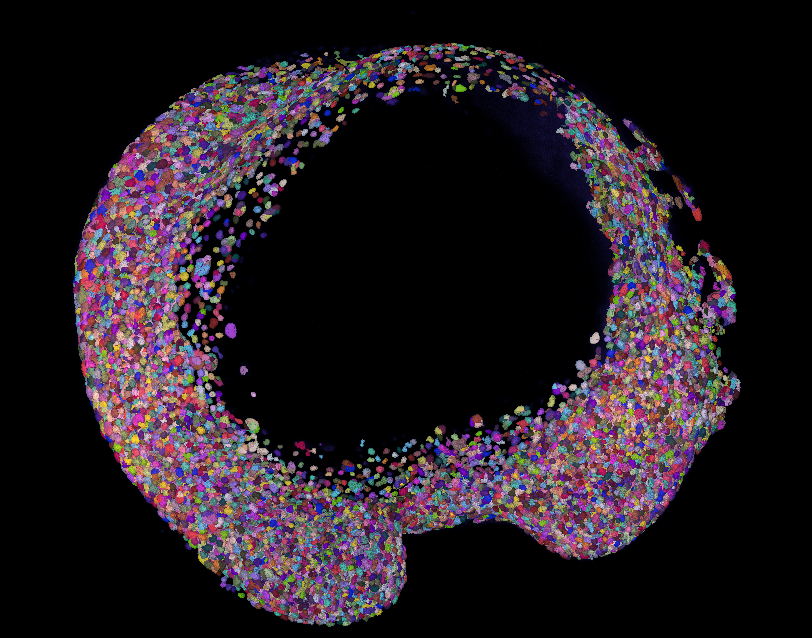}
        \caption{}
    \end{subfigure}
    \hfill
    \begin{subfigure}[b]{0.23\columnwidth}
        \includegraphics[width=\textwidth]{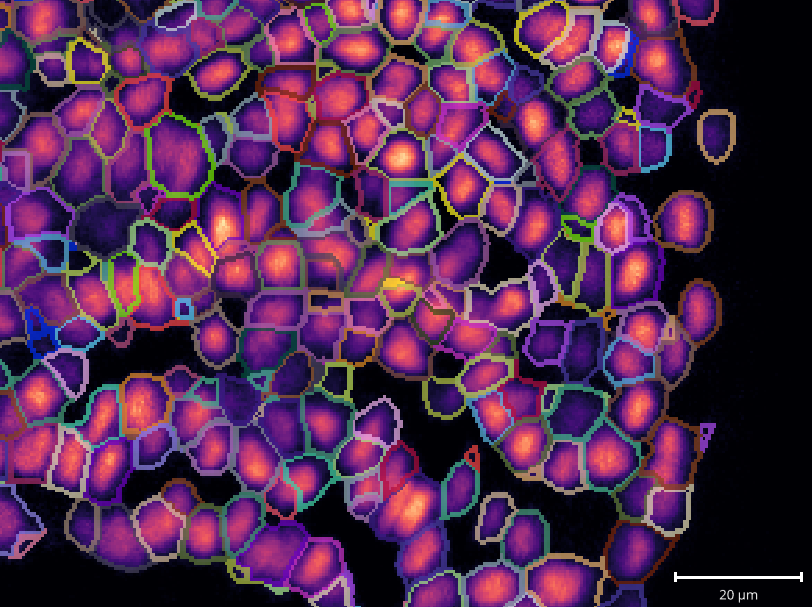}
        \caption{}
    \end{subfigure}
    \hfill
    \begin{subfigure}[b]{0.23\columnwidth}
        \includegraphics[width=\textwidth]{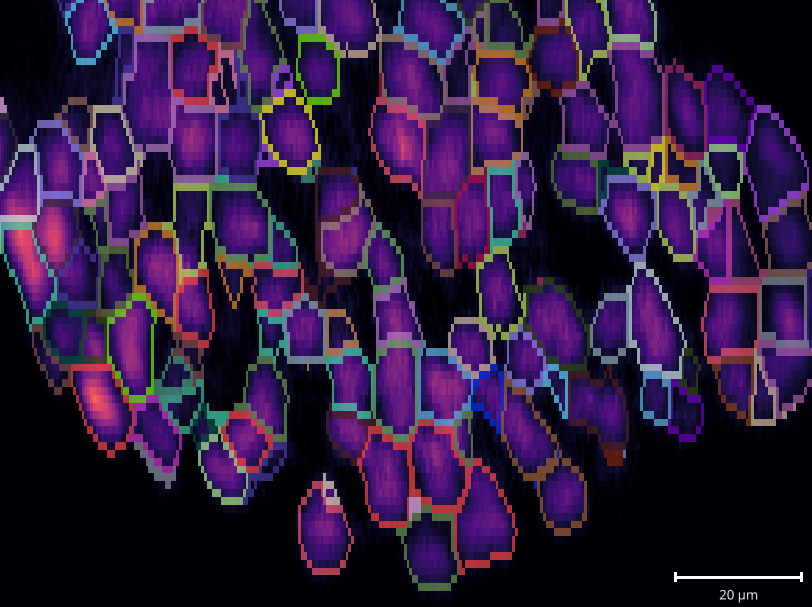}
        \caption{}
    \end{subfigure}
    \caption{Whole embryo segmentation results: (a) Maximum intensity projection; (b) 3D rendering of segmentation labels; (c) YX slice; and (d) ZY slice.}
    \label{f.wholeembryo}
\end{figure}


\section{Limitations \& Social Impact}

While the methodology is more robust to segmentation mistakes than non-joint segmentation and tracking methods, the space of possible segmentations is constrained from the input contour map. Therefore, contour maps that poorly corresponds to the cell boundaries greatly impar our methods performance.

The negative social impact of this work would be the employment of this method to assist unethical/immoral biological research, as with any other bioinformatics tool.

\section{Conclusion}
\label{s.conclusion}

In this paper, we introduced a novel approach to cell tracking in microscopy data that selects cell segments from hierarchical segmentations. Our method demonstrated state-of-the-art performance on three out of five datasets from the Cell Tracking Challenge~\cite{Ulman:2017:CellTrackingChallenge} and on the Epithelial Cell Benchmark~\cite{Funke:2018:BenchmarkEpithelialCellTracking}. To date, it is the only method to excel in both nuclei and membrane-based cell tracking while also scaling to terabyte-scale datasets.

However, the benefits of using contour maps for cell tracking extend beyond surpassing benchmark metrics and improving segmentation performance. The contour (\ie affinity) space offers more flexibility than the traditional domain of discrete labels, enabling the use of an ensemble of segmentation models and differential models for edge detection.

Therefore, the outlined goal of developing an effective joint segmentation and tracking method that supports a variety of inputs was achieved.

Overall, we believe our approach has significant potential for advancing cell tracking in complex microscopy data and opening new avenues for exploration in bioimaging research.
In future work, a model could be used to predict the ILP weights.

\section*{Acknowledgements}

We thank Ahmed Abbas, Ahmet Can Solak, Alexandre Falc\~ao, Guillaume Le Treut, Ilan Francisco, Laurent Najman, Paul Swoboda, and Sandy Schmid for discussions regarding this work and Chan Zuckerberg Biohub for the support.

%
%
\bibliographystyle{splncs04}
\bibliography{references}
\end{document}